\documentclass[letterpaper, 10 pt, conference]{ieeeconf}

\IEEEoverridecommandlockouts

\usepackage{graphics}
\usepackage{graphicx}
\usepackage{amssymb}
\usepackage{subfigure}
\usepackage{caption}
\usepackage{amsmath} 
\usepackage{booktabs}
\usepackage{tabularx}

\title{ \LARGE \bf Experience-Driven Continual Learning of Terrain Traversability for Quadruped Robots}

\author{Luca Bricarello\textsuperscript{1,2},
João Carlos Virgolino Soares\textsuperscript{1},
Alberto Sanchez-Delgado\textsuperscript{1},\\[0.4em]
Fulvio Mastrogiovanni\textsuperscript{2},
Claudio Semini\textsuperscript{1}%
\thanks{\textsuperscript{1} Dynamic Legged Systems Lab,
Istituto Italiano di Tecnologia, Genoa, Italy.
\quad\textsuperscript{2} Università degli Studi di Genova, Genoa, Italy.
Correspondence: {\tt\small joao.virgolino@iit.it}}}

\begin{document}
\raggedbottom

\maketitle

\begin{abstract}

Safe and efficient quadruped navigation over unfamiliar terrain requires predicting terrain--robot interaction before contact: geometry and visual appearance alone cannot reveal how the robot will slip, load its feet, or expend energy.
This paper presents a continual learning pipeline that uses locomotion experience to learn these interaction outcomes from pre-contact images and continually updates the predictions as new contacts are observed.
Pre-contact descriptors, produced by a DINOv3 backbone model frozen during training, are mapped to five proprioceptive indicators weighted according to measurement reliability: planar foot slip, mean normal ground-reaction force, traction index, cost of transport, and touchdown loading rate.
A compact evidential regressor allows us to predict these indicators together with aleatoric and epistemic uncertainty from the visual descriptors.
Continual adaptation combines bounded experience replay with a validation gate: candidate models replace the deployed predictor only when they improve performance on recent held-out data while keeping degradation on historical held-out data within a prescribed tolerance.
Predictions and epistemic uncertainty are projected into a local multilayer map and combined into a conservative traversability score map whose property weights can be adjusted without retraining.
The resulting map is used for downstream navigation tests.
The ROS~2 implementation supports evaluation on a Unitree Go2 in simulation and on hardware, with models trained separately in each domain.
On a sequential hardware stream over three previously unseen terrains, gated replay reduces final anchor negative log-likelihood (NLL) degradation by 23.1\% relative to replay without the gate while attaining similar new-terrain adaptation.

\end{abstract}

\begin{figure}[t]
    \centering
    \includegraphics[width=1.0\linewidth]{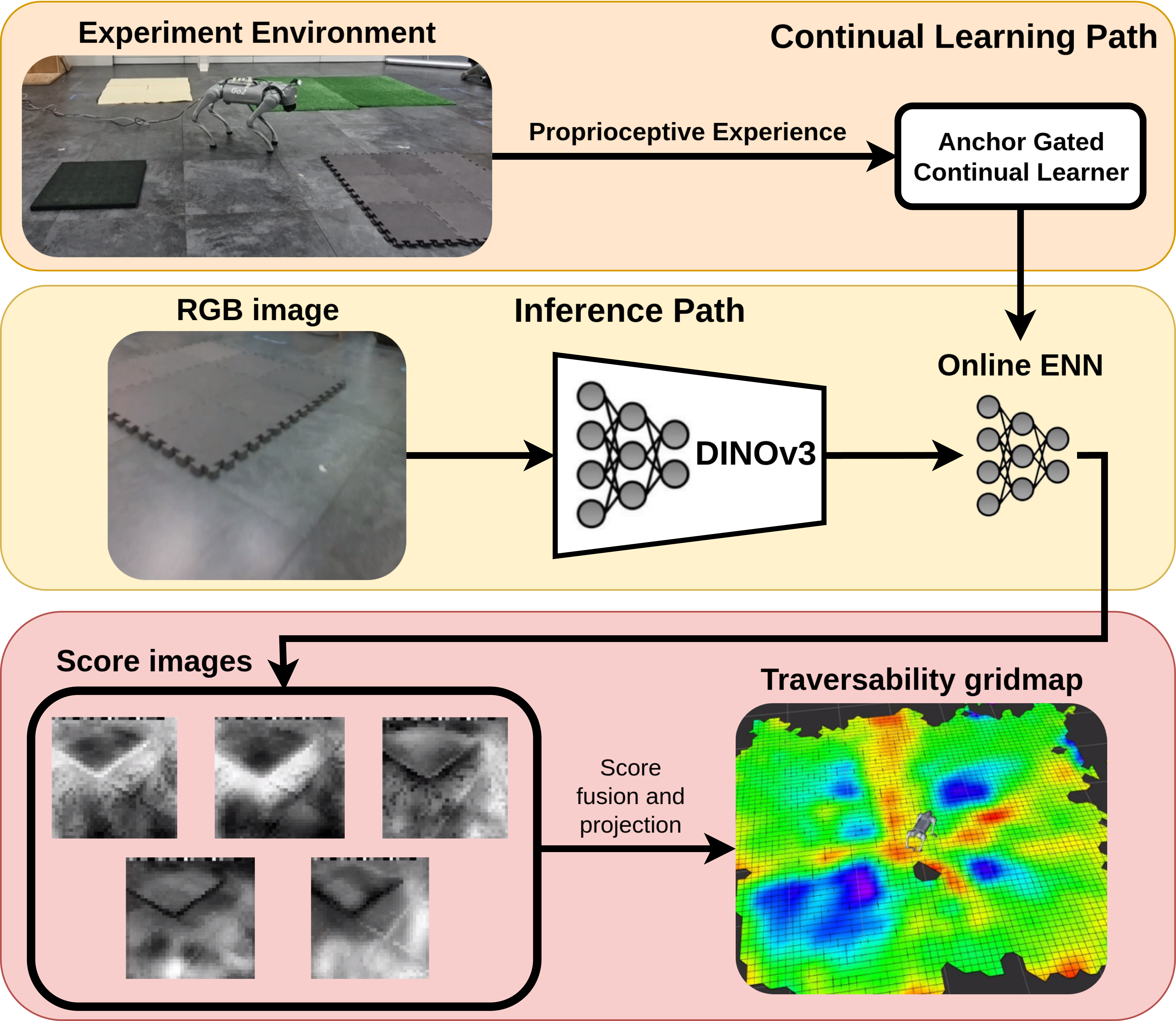}
    \caption{Qualitative hardware dataflow: RGB features are extracted by frozen DINOv3, converted by the evidential regressor into five interaction-property score images, and projected into a traversability grid map. The foam surfaces receive lower traversability scores. Artificial grass scores close to the laboratory floor, but slightly lower because it is perceived as more slippery.}

    \label{fig:dataflow}
\end{figure}

%%%%%%%%%%%%%%%%%%%%%%%%%%%%%%%%%%%%%%%%%%%%%%%%%%%%%%%%%%%%%%%%%%%%%%%%%%%%%%%%
\section{INTRODUCTION}

Planning where a quadruped robot can step and how the terrain will respond to loading are equally important navigation considerations. 
Terrain elevation maps enable foothold and path planning on rough terrain~\cite{Wermelinger2016NavigationPF,Miki2022ElevationMF}, yet identical terrain geometry can obscure variations in friction, compliance, or load-bearing properties~\cite{Chen2024IdentifyingTP,Medeiros2025LoadBearingAF}. 
Vegetation perceived as blocking a path may be pushed through instead ~\cite{sathyamoorthy2023vernvegetationawarerobotnavigation,Li2023SeeingTT}.
Thus, terrain traversability depends on terrain-robot interaction effects rather than terrain appearance alone.
We aim to leverage information about terrain ahead of the robot to anticipate these interaction effects. 

Vision and proprioception are two sources of information about terrain-robot interactions. Optical images capture information about terrain before it is contacted, and internal sensors report how the terrain responds to attempted locomotion~\cite{Elnoor2023ProNavPT,Kang2023StateEA}.
 
Self-supervised approaches relate these two data streams by registering post-contact state to corresponding images~\cite{Wellhausen2019WhereSI,Chen2024IdentifyingTP}.
Features learned from visuals can then be used to rapidly adapt predictions with small amounts of robot experience ~\cite{Mattamala2024WildVN,Sivaprakasam2024SALONSA}. 
Yet robot adaptation methods must also be able to handle continued accumulation of experience: online fine-tuning may forget past terrain encounters~\cite{Ma2024IMOSTIM,Hindel2026SelfSupervisedOR}, while models may become overconfident on out-of-distribution observations ~\cite{Wellhausen2020SafeRN,Cai2023EVORADE}. 

Prior work has addressed parts of these challenges through visual prediction of physical properties~\cite{Wellhausen2019WhereSI,Chen2024IdentifyingTP}, retention of robot experience during adaptation~\cite{Ma2024IMOSTIM,Hindel2026SelfSupervisedOR}, and uncertainty-aware navigation costs~\cite{Sivaprakasam2024SALONSA,Cai2023EVORADE}.
This paper presents a quadruped traversability pipeline that combines predictions of the robot's own interaction outcomes with continual adaptation and uncertainty-aware mapping.
It associates pre-contact visual descriptions with five confidence-weighted foot-contact indicators: planar foot slip, mean normal ground-reaction force, traction index, cost of transport, and touchdown loading rate. A compact evidential regressor predicts each indicator and its uncertainty from frozen visual features, allowing the source of terrain difficulty to be examined separately for each property.
After offline training, the regressor adapts with new and replayed contacts.
We call the candidate acceptance test a \emph{validation gate}: an update replaces the deployed predictor only if it improves negative log-likelihood (NLL) on recent held-out contacts without exceeding a prescribed tolerance for degradation on historical held-out contacts.

We hypothesize that adding historical validation to recent-validation replay improves retention without impairing adaptation to new terrain.

To the best of the authors' knowledge, this is the first quadruped visual traversability system to combine confidence-weighted prediction of multiple foot-contact outcomes with validation-gated continual adaptation on both recent and historical experience.
The main contributions of this work are:

\begin{itemize}
    \item A novel validation-gated continual learning strategy that combines replay-based training with held-out recent and historical validation. In a one-stream hardware ablation, the historical gate reduced final anchor NLL degradation by 23.1\% relative to replay with recent validation alone, with similar new-terrain adaptation.
    \item A contact-level supervision formulation that associates pre-contact visual observations with five proprioceptive interaction indicators, weighted by their individual measurement confidences.
    \item A property-resolved evidential traversability map that preserves individual interaction predictions and their epistemic uncertainties, enabling adjustment of navigation preferences and uncertainty penalties without retraining the predictor.
\end{itemize}

\section{RELATED WORK}

\subsection{Geometric and Semantic Traversability}

Geometric approaches evaluate terrain morphology against the robot's locomotion skills. 
Wermelinger et al.~\cite{Wermelinger2016NavigationPF} integrate geometric traversability assessment with navigation planning for a legged robot.
Separately, Miki et al.~\cite{Miki2022ElevationMF} accelerate elevation-map construction with GPU processing and demonstrate the map in quadrupedal locomotion experiments.
Geometric maps alone may not distinguish vegetation from the underlying support surface or determine whether a surface can sustain contact.
VERN~\cite{sathyamoorthy2023vernvegetationawarerobotnavigation} incorporates vegetation properties into costmap construction, while Li et al.~\cite{Li2023SeeingTT} use foothold supervision to recover the support surface beneath grass.

Human demonstrations offer another source of visual terrain knowledge.
Kim et al.~\cite{Kim2024LearningST} generate semantic traversability annotations from egocentric walking videos and deploy the predictor on a quadruped.
STEPP~\cite{gidius2025WatchYS} learns the distribution of pose-projected visual features from human walking and detects unfamiliar terrain by reconstruction error. 
These methods reduce annotation effort, but the demonstrated traversable regions do not directly measure the target robot's planar foot slip, mean normal ground-reaction force, or cost of transport.

\subsection{Proprioceptive Assessment and Physical Properties}

ProNav~\cite{Elnoor2023ProNavPT} uses joint, force, and current signals to assess terrain-induced instability and resistance to locomotion, including risks of slipping or leg entrapment.
Kang et al.~\cite{Kang2023StateEA} account for slip in state estimation and incorporate stiffness into terrain maps. 
Sanchez-Delgado et al.~\cite{SnchezDelgado2025TowardsPT} construct proprioceptive maps of elevation, slip, energy, and stability under lunar gravity in simulation. 
Dedicated probing supplies additional evidence: Haddeler et al.~\cite{Haddeler2022TraversabilityAW} combine vision with a force-sensing arm to assess collapsibility, while Medeiros et al.~\cite{Medeiros2025LoadBearingAF} use quadruped joint measurements and planned foot probing motions to assess load-bearing capacity.

To anticipate contact, Wellhausen et al.~\cite{Wellhausen2019WhereSI} transfer force-derived foothold labels into images. 
Chen et al.~\cite{Chen2024IdentifyingTP} train a physical decoder in simulation to supervise visual friction and stiffness prediction. 
Ewen et al.~\cite{Ewen2022TheseMA} instead infer friction distributions through a Bayesian semantic map with class-dependent property priors. 
A remaining gap is to connect several effects measured during the robot's own foot contacts to what it saw before contact, while accounting for the different reliability of each measurement.
To address this gap, our formulation associates each completed stance with five interaction indicators and their individual measurement confidences.
These confidence-weighted targets supervise predictions from pre-contact visual observations, capturing the robot's experienced response under the observed operating conditions.

\subsection{Learning from Experience and Continual Adaptation}

STERLING~\cite{Karnan2023SelfSupervisedTR} learns terrain representations by aligning visual and non-visual experience. 
De Miguel et al.~\cite{Miguel2025IMT} cluster encoded elevation--texture patches and assess new terrain by similarity to accumulated traversals. 
WVN~\cite{Mattamala2024WildVN} learns a visual traversability head online using frozen features and robot-generated supervision. 
SALON~\cite{Sivaprakasam2024SALONSA} adapts cost and speed maps from experienced roughness, using Gaussian-process prediction and a managed experience buffer. 
Multimodal navigation also benefits from feedback: AMCO~\cite{Elnoor2024AMCOAM} combines visual and proprioceptive costmaps according to image reliability, and TOP-Nav~\cite{Ren2024TOPNavLN} corrects visual estimates using locomotion evaluations.

Retention during sequential learning is addressed explicitly by IMOST~\cite{Ma2024IMOSTIM}, which combines scene-aware incremental memory with self-supervised segmentation, and by COTRATE~\cite{Hindel2026SelfSupervisedOR}, which learns proprioceptive assessments and retains diverse visual features in compact replay. 
Lee et al.~\cite{Lee2025ContinualLF} evaluate continual traversability learning on a skid-steering wheeled robot. Their predictor combines visual and geometric terrain features with robot velocity to predict linear and angular traction using a probabilistic ensemble that models aleatoric and epistemic uncertainty.
Their continual learning framework uses generative experience recall to retain prior knowledge without storing past datasets, and accounts for the uncertainty of recalled samples when updating the recall model.
These approaches address retention through the selection, replay, or generation of past experience.
Our approach additionally separates replay-based optimization from candidate acceptance: recent and historical held-out data determine whether an update replaces the deployed model.
Acceptance requires improved predictive likelihood on recent experience and limits historical degradation relative to the current model.

\subsection{Uncertainty and the Navigation Interface}

Unfamiliarity and poor traversability are related but distinct.
Wellhausen et al.~\cite{Wellhausen2020SafeRN} detect anomalous terrain from multimodal observations, while STEPP~\cite{gidius2025WatchYS} uses reconstruction error as a familiarity measure. 
Distributional approaches expose uncertainty more directly: Ewen et al.~\cite{Ewen2022TheseMA} maintain property distributions, SALON~\cite{Sivaprakasam2024SALONSA} adjusts costs and speeds using predictive variance, and EVORA~\cite{Cai2023EVORADE} combines evidential traction distributions with feature-space epistemic uncertainty and risk-aware planning.

A complementary planning formulation predicts action outcomes rather than cell costs. 
Kim et al.~\cite{Kim2022LearningFD} combine a learned forward dynamics
model with trajectory sampling for quadruped navigation; Roth et al.~\cite{Roth2025LearnedPF} predict future states and failure probabilities from geometry and proprioceptive history. 
Our representation builds on deep evidential regression~\cite{Amini2019DeepER} to estimate individual interaction outcomes and their uncertainties.
The spatial map retains separate property and epistemic uncertainty layers before aggregation into a traversability score.
This separation allows navigation preferences and uncertainty penalties to be adjusted while keeping the learned predictor fixed.

Table~\ref{tab:related_work_comparison} compares the approaches most relevant to this work by predicted outputs, online adaptation, retention, uncertainty, and update gating.

\begin{table*}[t]
    \centering
    \caption{Comparison of selected traversability methods.
    \textit{Online} denotes predictor adaptation from newly acquired deployment experience.
    \textit{Gate} denotes candidate acceptance based jointly on improved predictive performance on recent held-out data and a prescribed tolerance on historical held-out degradation relative to the deployed model.
    A/E: aleatoric/epistemic uncertainty; --: not reported as a component of the described method.}
    \label{tab:related_work_comparison}
    \small
    \setlength{\tabcolsep}{4pt}
    \renewcommand{\arraystretch}{1.18}
    \begin{tabularx}{\textwidth}{@{}
        >{\raggedright\arraybackslash}p{0.12\textwidth}
        >{\raggedright\arraybackslash}X
        c
        >{\raggedright\arraybackslash}p{0.22\textwidth}
        >{\raggedright\arraybackslash}p{0.25\textwidth}
        c@{}}
        \toprule
        \textbf{Method} & \textbf{Predicted output} & \textbf{Online}
        & \textbf{Memory / retention} & \textbf{Uncertainty / confidence} & \textbf{Gate} \\
        \midrule
        WVN~\cite{Mattamala2024WildVN}
        & Traversability score & Yes
        & Mission graph & Reconstruction-based confidence & -- \\
        Chen et al.~\cite{Chen2024IdentifyingTP}
        & Friction, stiffness & Yes
        & Mission graph & Anomaly-based confidence mask & -- \\
        EVORA~\cite{Cai2023EVORADE}
        & Linear/angular traction & --
        & -- & Evidential distributions + feature density (A/E) & -- \\
        SALON~\cite{Sivaprakasam2024SALONSA}
        & Roughness, speed & Yes
        & Experience buffer & GP predictive variance & -- \\
        IMOST~\cite{Ma2024IMOSTIM}
        & Traversability & Yes
        & Incremental memory & Reconstruction-based anomaly / sampling scores & -- \\
        Lee et al.~\cite{Lee2025ContinualLF}
        & Linear/angular traction & Yes
        & Generative recall & Probabilistic ensemble (A/E) & -- \\
        COTRATE~\cite{Hindel2026SelfSupervisedOR}
        & Traversability score & Yes
        & Selective feature replay & -- & -- \\
        \textbf{Proposed}
        & \textbf{Five interaction indicators} & \textbf{Yes}
        & \textbf{Bounded replay} & \textbf{Evidential regression (A/E)} & \textbf{Yes} \\
        \bottomrule
    \end{tabularx}
\end{table*}

\section{METHODOLOGY}
\label{sec:methodology}

\subsection{Problem Formulation and System Overview}

We define traversability as a robot-dependent outcome of interaction with the terrain, rather than a semantic terrain category.
Accordingly, instead of assigning semantic labels such as sand or gravel, the proposed system predicts the expected interaction outcomes when the robot traverses visually observed terrain. 
We represent these outcomes through five complementary indicators, defined below.
These predictions are subsequently converted into property-specific traversability scores and fused into a navigation cost representation.
Fig.~\ref{fig:dataflow} illustrates this prediction-to-map process on hardware.

At time $t$, an RGB image $I_t$ is processed by a frozen DINOv3 ViT-S/16 backbone~\cite{Simoni2025DINOv3}.
Its patch tokens form a spatial feature map $\mathbf{X}_t\in\mathbb{R}^{36\times36\times384}$, where $\mathbf{x}_{t,i,j}\in\mathbb{R}^{384}$ describes one image region.
For each completed foot contact $k$, proprioception provides the multi-property target

\begin{equation}
    \mathbf{y}_k =
    \left[
    d_{\mathrm{slip}},\ \overline{F}_{n},\ r_{\mathrm{tr}},\
    \mathrm{CoT},\ \dot{F}^{\mathrm{td}}_{n}
    \right]^{\mathsf T}_{k},
    \label{eq:paper_property_vector}
\end{equation}

together with property-specific confidence values $\mathbf{c}_k\in[0,1]^5$. The five quantities respectively describe planar foot slip, mean normal ground-reaction force, traction index, cost of transport, and touchdown loading rate. We abbreviate ground-reaction force as GRF and cost of transport as CoT below. Their equations are given in Sec.~\ref{sec:paper_contact_indicators}.

The central requirement is that supervision remains causal. 
Let $t_k^{\mathrm{td}}$ and $t_k^{\mathrm{lo}}$ denote touchdown and liftoff, respectively.
A training sample is generated only from a visual observation acquired before contact,

\begin{equation}
    \mathcal{S}_k=\left(\mathbf{x}_{t_v,i,j},\mathbf{y}_k,\mathbf{c}_k\right),
    \qquad t_v\leq t_k^{\mathrm{td}}<t_k^{\mathrm{lo}}.
    \label{eq:paper_causal_sample}
\end{equation}

\begin{figure*}[t]
    \centering
    \includegraphics[width=0.94\linewidth]{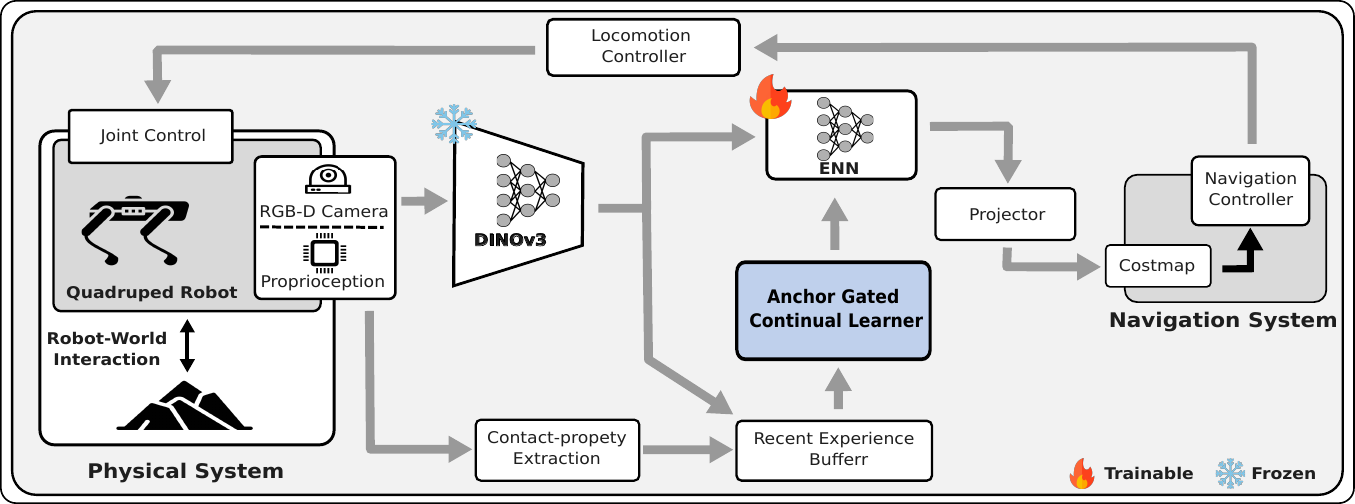}
    \caption{Overview of the proposed experience-driven traversability pipeline. The deployed model performs uninterrupted inference while a separate candidate is trained from newly acquired visual--proprioceptive experience.}
    \label{fig:paper_pipeline}
\end{figure*}

The resulting samples can either be stored for offline training or used immediately by the continual learner to update the current model. During deployment, the same visual features are processed by an evidential regressor and projected into a local multi-layer traversability map. 
Fig.~\ref{fig:paper_pipeline} summarizes the inference and experience-learning paths.

\subsection{Causal Visual--Proprioceptive Supervision}
\label{sec:paper_contact_indicators}

Each RGB frame is evaluated by the DINOv3 visual backbone, which remains frozen during both offline and online learning.
Freezing it reduces the onboard training cost and, crucially, keeps the representation stationary: descriptors stored at different times remain compatible with the same lightweight downstream model, which instead can be updated with continual learning.
The input image is resized to $576\times576$ pixels, from which DINOv3 extracts a \(36 \times 36\) grid of local tokens.
We use the ROS~2 \texttt{odom} frame as the common local odometry coordinate system for visual patches, foot-contact positions, and maps; superscript $o$ denotes quantities expressed in this frame.
Registered depth and camera pose are evaluated at the original RGB timestamp and used to project each valid patch into a $5$~cm contact-association grid in this frame.
We approximate its ground footprint by deprojecting the four patch corners at the median depth of the corresponding image region.

In parallel, each foot is tracked independently from touchdown to liftoff to compute the five indicators listed above.
Let ${}^{o}\mathbf{p}_{f,k}(t)$ be the odometry-frame position of the contacting foot in contact $k$. Planar foot slip is its horizontal displacement during stance,

\begin{equation}
    d_{\mathrm{slip},k}=
    \left\|{}^{o}\mathbf{p}_{f,k}(t_k^{\mathrm{lo}})-
    {}^{o}\mathbf{p}_{f,k}(t_k^{\mathrm{td}})\right\|_{xy}.
    \label{eq:paper_planar_foot_slip}
\end{equation}

Contact-force indicators use joint torque measurements and the analytic leg Jacobian. At each synchronized stance sample $t_r$, the GRF is estimated with a damped inverse to remain well-conditioned near kinematic singularities,

\begin{equation}
    \begin{aligned}
        {}^{b}\widehat{\mathbf{F}}_k(t_r)
        &= \mathbf{J}_k(t_r)\\
        &\quad\left(\mathbf{J}_k^{\mathsf T}(t_r)\mathbf{J}_k(t_r)
        +\mu^2\mathbf{I}\right)^{-1}\boldsymbol{\tau}_k(t_r).
    \end{aligned}
    \label{eq:paper_damped_grf}
\end{equation}
We use $\mu=10^{-3}$.

We estimate the local support plane from the odometry-frame positions of at least three feet simultaneously in stance. When fewer feet are in contact or the plane estimate is degenerate, we use the horizontal plane.
The estimated force is rotated into the odometry frame and resolved along the upward-pointing normal of this plane into the normal component $F_{n,k}(t_r)$ and tangential component $\mathbf{F}_{t,k}(t_r)$. Let $N_{F,k}$ be the number of valid force samples during stance. Mean normal ground-reaction force and traction index are

\begin{align}
    \overline{F}_{n,k} &=
    \frac{1}{N_{F,k}}\sum_{r=1}^{N_{F,k}}F_{n,k}(t_r),
    \label{eq:paper_mean_normal_grf}\\
    r_{\mathrm{tr},k} &=
    \frac{\overline{\|\mathbf{F}_{t,k}\|}}
    {|\overline{F}_{n,k}|+\epsilon_F}.
    \label{eq:paper_traction_index}
\end{align}

Here $\overline{\|\mathbf{F}_{t,k}\|}$ is the mean tangential-force magnitude over the same samples, and $\epsilon_F=1$~N regularizes small normal loads.
The traction index measures the tangential-force magnitude sustained relative to normal load during the observed gait; it is not a direct estimate of the terrain friction coefficient. In our scoring convention, larger values of this raw index receive a higher traction score, while planar foot slip separately penalizes loss of grip.

Let $\overline{P}_{\mathrm{abs},k}$ be the mean absolute joint mechanical power during stance, where $P_{\mathrm{abs}}(t)=\sum_j|\tau_j(t)\dot q_j(t)|$, and let $\overline{v}_{xy,k}$ be the mean planar base speed in the same interval. For the touchdown window $[t_k^{\mathrm{td}},t_k^{\mathrm{td}}+50\,\mathrm{ms}]$, let $t_a<t_b$ be the first and last valid normal-force sample times. Cost of transport and touchdown loading rate are

\begin{align}
    \mathrm{CoT}_k &=
    \frac{\overline{P}_{\mathrm{abs},k}}
    {mg\overline{v}_{xy,k}},
    \label{eq:paper_cost_of_transport}\\
    \dot{F}^{\mathrm{td}}_{n,k} &=
    \frac{F_{n,k}(t_b)-F_{n,k}(t_a)}{t_b-t_a}.
    \label{eq:paper_touchdown_loading_rate}
\end{align}

Here $m$ is the robot mass and $g$ the gravity constant. Touchdown loading rate requires at least two valid force samples in the $50$~ms window.

Let $N_k$ be the number of synchronized samples collected during stance, $\bar{\kappa}_k$ the mean Jacobian conditioning ratio $\sigma_{\min}(\mathbf J)/\sigma_{\max}(\mathbf J)$, and $N_k^{\mathrm{td}}$ the number of force samples in the touchdown window.
The property confidences are

\begin{align}
c_k^{\mathrm{slip}} &=
 \min(1,N_k/N_{\min}), \notag\\
c_k^{F_n} &=
 \min(1,\bar{\kappa}_k/\kappa_{\min}), \notag\\
c_k^{\mathrm{tr}} &=
c_k^{F_n}\min(1,|\bar F_{n,k}|/(10\epsilon_F)), \notag\\
c_k^{\mathrm{CoT}} &=
 \mathbb{I}[\bar v_{xy,k}\geq v_{\min}], \notag\\
c_k^{\mathrm{load}} &=
 \mathbb{I}[N_k^{\mathrm{td}}\geq2]
 \min(1,N_k^{\mathrm{td}}/3)
 \min(1,\bar{\kappa}_k^{\mathrm{td}}/\kappa_{\min}).
\end{align}

Missing measurements yield zero confidence.

A contact is rejected when the minimum property confidence is below $0.25$; otherwise the individual confidences weight the corresponding learning objectives.

The full property vector becomes available only at liftoff, although it refers to the contact location recorded at touchdown. 
The previously saved grid map associates the five computed indicators with visual descriptors at the foot-contact position.
For a touchdown location, the system selects the most recent feature map that (i) precedes touchdown, (ii) is at most $5$~s old, and (iii) contains a projected patch footprint covering the same $5$~cm contact-association cell. These cells match contacts to visual patches; the output traversability map instead uses $0.1$~m cells, as described in Sec.~\ref{sec:paper_spatial_map}.
This procedure associates the outcome of a step with terrain that was actually visible before interaction and prevents future observations from leaking into the training sample.

\subsection{Uncertainty-Aware Multi-Property Regression}

The predictor is a compact multi-head evidential MLP based on deep evidential regression~\cite{Amini2019DeepER}. 
Standardized $384$-dimensional visual descriptors are processed by a shared $384\!\rightarrow\!128\!\rightarrow\!64$ trunk, followed by five independent $64\!\rightarrow\!32\!\rightarrow\!4$ heads.
Each head outputs the parameters $(\gamma,\nu,\alpha,\beta)$ of a Normal--Inverse-Gamma distribution (NIG). We enforce $\nu>0$, $\alpha>1$, and $\beta>0$ using softplus transformations. 
The predictive mean is $\gamma$, whereas aleatoric and epistemic uncertainty are

\begin{equation}
    \sigma^2_{\mathrm{ale}}=\frac{\beta}{\alpha-1},
    \qquad
    \sigma^2_{\mathrm{epi}}=\frac{\beta}{\nu(\alpha-1)}.
    \label{eq:paper_evidential_uncertainty}
\end{equation}

This provides the five predictions of the considered properties and their uncertainties in a single forward pass. 
Only epistemic uncertainty is propagated to the traversability map, while aleatoric uncertainty is used only for model evaluation.

For a mini-batch $\mathcal{B}$, the network is trained with a confidence-weighted evidential objective

\begin{equation}
    \begin{aligned}
    \ell_{n,p}&=\mathcal{L}^{\mathrm{NIG}}_{n,p}
    +\lambda_e|\widetilde y_{n,p}-\gamma_{n,p}|
    (2\nu_{n,p}+\alpha_{n,p}),\\
    \mathcal{L}_{\mathcal B}
    &=\frac{\sum_{n\in\mathcal B}\sum_p c_{n,p}\ell_{n,p}}
    {\sum_{n\in\mathcal B}\sum_p c_{n,p}} .
    \end{aligned}
    \label{eq:paper_evidential_loss}
\end{equation}

\noindent where $\mathcal{L}^{\mathrm{NIG}}$ is the negative log-likelihood (NLL) of the Student-$t$ distribution induced by the NIG parameters. 
The objective of the second term is to discourage confident but inaccurate predictions. 
Input and target scalers are fitted only on the offline training partition and then frozen.
Samples sharing a timestamp or an identical descriptor are grouped before the offline train/validation/test split to prevent contact-correlated observations from leaking across partitions.

At inference, predicted means and epistemic variances are transformed back to their respective physical units. 
Before normalization, we orient the raw predictions with the sign vector $(-1,+1,+1,-1,+1)$ in the order of Eq.~\ref{eq:paper_property_vector}. Thus planar foot slip and cost of transport are inverted, while mean normal ground-reaction force, traction index, and touchdown loading rate retain their signs under our operational scoring convention. Larger oriented values, and hence larger normalized property scores, denote better traversability.
Each property is normalized between $[0,1]$ using fixed 5th and 95th percentiles computed on reference predictions, thus building a score for each property. 
The variances are scaled by the square of the same interval, but are not clipped, preserving large uncertainty on out-of-distribution terrain.

\subsection{Guarded Continual Adaptation}

During locomotion, accepted samples are assigned deterministically to recent training and validation buffers using a hash of the session identifier and touchdown timestamp. This keeps simultaneous contacts in the same partition. 
Then, nearly identical descriptors are rejected using cosine similarity, avoiding an artificial over-representation of visually uniform terrain. 
A nominal update starts after the collection of $50$ recent training samples and at least $10$ held-out samples.

To reduce catastrophic forgetting, each recent block is combined with a bounded replay set,

\begin{equation}
    \mathcal{D}^{\mathrm{train}}_k =
    \mathcal{D}^{\mathrm{rec}}_k\cup\mathcal{D}^{\mathrm{rep}}_k,
    \qquad |\mathcal{D}^{\mathrm{rep}}_k|\leq150.
    \label{eq:paper_replay_set}
\end{equation}

Replay draws from previously consolidated online samples and, when available, samples collected offline in previous experiments. 
The online memory is bounded to $1000$ elements and sampled across planar foot slip quantiles so that rare contacts with high planar foot slip are not displaced by nominal contacts. 
Replay data is used in gradient updates, whereas a separate anchor set remains strictly held out and measures retention of prior knowledge.

At update $k$, a candidate is initialized from the deployed parameters $\boldsymbol{\theta}^{(k)}$ and optimized in a background thread. 
Let $J_{\mathrm{rec}}$ and $J_{\mathrm{anc}}$ denote confidence-weighted NIG negative log-likelihood on the recent validation set and the anchor, respectively. 
Promotion requires

\begin{align}
    J_{\mathrm{rec}}(\boldsymbol{\theta}^{\mathrm{cand}})
    &<J_{\mathrm{rec}}(\boldsymbol{\theta}^{(k)})-\delta_{\min},
    \notag\\
    J_{\mathrm{anc}}(\boldsymbol{\theta}^{\mathrm{cand}})
    &\leq J_{\mathrm{anc}}(\boldsymbol{\theta}^{(k)}) \notag\\
    &\quad+\rho_{\max}\max\!\left(
    |J_{\mathrm{anc}}(\boldsymbol{\theta}^{(k)})|,10^{-6}\right),
    \label{eq:paper_promotion_gate}
\end{align}

In the sequential hardware experiment we set $\delta_{\min}=0$ and $\rho_{\max}=0.0015$; thus, an update must strictly improve performance on recent experience while allowing at most a $0.15\%$ relative increase in anchor NLL at each promotion opportunity. 
Candidates producing non-finite outputs are also rejected. 
A successful checkpoint is written and validated before atomically replacing the inference model; otherwise the previous model keeps producing predictions without interruption.

Continual adaptation starts from an offline-trained evidential checkpoint.
A partition excluded from checkpoint fitting and model selection supplies the initial historical anchor.

\subsection{Spatial Traversability Mapping}
\label{sec:paper_spatial_map}

As described above, the regressor produces five normalized means and epistemic variances for each image patch. 
Synchronized depth and camera pose are then used to project each predicted patch footprint into a $15\times15$~m robot-centered GridMap with $0.1$~m resolution in the odom frame. 
When multiple patches overlap a cell $c$ in the same frame, their first and second moments are aggregated as

\begin{equation}
    \overline{\mu}_{c,p}=\frac{1}{M_c}\sum_j\mu_{j,p},\qquad
    \overline{v}_{c,p}=\frac{1}{M_c}\sum_j
    (v_{j,p}+\mu_{j,p}^{2})-\overline{\mu}_{c,p}^{2}.
    \label{eq:paper_spatial_moments}
\end{equation}

The resulting variance therefore contains both predicted epistemic uncertainty and disagreement among overlapping patches.
One temporal exponential update with $\alpha_{\mathrm{EMA}}=0.05$ is then applied per observed cell and frame.

Before fusion, each property is conservatively corrected as $q_{c,p}=\mu_{c,p}-\lambda\sqrt{v_{c,p}}$, with $\lambda=0.15$. 
The final traversability is then computed as a normalized weighted sum

\begin{equation}
    \tau_c=\operatorname{clip}\!\left(
    \frac{\sum_{p\in\mathcal A}w_p q_{c,p}}
         {\sum_{p\in\mathcal A}w_p},0,1\right),
    \qquad \mathcal A=\{p\mid w_p>0\}.
    \label{eq:paper_traversability_fusion}
\end{equation}

The weights are configurable because the informativeness of each interaction property depends on the domain in which the robot is operating. 
For example, properties affected by contact compliance can be disabled in rigid-contact simulation, while all five remain available for hardware experiments. 
The map keeps the individual property and uncertainty layers in addition to $\tau_c$, so changing the navigation objective does not require retraining the predictor. 
Unobserved cells remain unknown rather than being implicitly marked as safe, and severe low-score conditions can be exposed as lethal cells to the downstream planner.

\section{RESULTS}
\label{sec:results}

\subsection{Experimental Setup and Protocol}
\label{sec:experimental_setup}

We evaluate property prediction, continual adaptation, and autonomous navigation on a Unitree Go2 quadruped robot in two scenarios.
Isaac Sim provides an environment containing three visually distinct surfaces, one with high friction (\(\mu_d = 1.1, \mu_s = 1.2\)) and two with lower friction (\(\mu_d = 0.45, \mu_s = 0.5\)).
Hardware property prediction and mapping use the laboratory floor, foam, textured foam, artificial grass, and a rubber patch; a separate continual-learning stream uses artificial grass, textured foam, and foam. Surface identities define the evaluation protocol but are not used as learning labels; updates exploit contact-derived supervision.
The difficulty of the hardware surfaces is established by independently measured interaction outcomes, rather than being assigned based on appearance.

The scoring in simulation employs planar foot slip and traction index, with $\mathbf{w}_{\mathrm{sim}}=[0.5 , 0 , 0.5 , 0 , 0]$, while the hardware uses $\mathbf{w}_{\mathrm{real}}=[0.4, 0.05, 0.1, 0.05, 0.3]$.
The simulation study evaluates and fuses planar foot slip and traction index only, because its rigid-contact model does not provide representative evidence for the broader interaction vector. The hardware study evaluates all five indicators.
On the real robot, we use a Unitree Go2 with an externally mounted RealSense D435 camera and the standard Go2 controller. The pipeline runs on an external PC with an RTX 5050 GPU, an Intel Core 7 CPU, and 16GB of RAM.
The models are trained separately for each domain; the hardware trials therefore evaluate the same pipeline, without assuming any model transfer from simulation to the real world.

The simulation and hardware datasets are disjoint and their regressors are trained independently.
Offline prediction and closed-loop navigation are assessed separately from the sequential hardware continual-learning comparison.
In that comparison, CL-P and Replay start from the same regressor trained only on laboratory-floor contacts and process the same recorded contact stream.
Both methods train with bounded online replay and select candidates using recent held-out contacts; CL-P additionally applies the historical anchor gate.
Navigation is evaluated in independent closed-loop simulation trials.

\subsection{Property Prediction and Calibration}

Initially, we evaluate Offline regressors trained on separate multi-terrain acquisition datasets for simulation and hardware.
These datasets contain observations from all terrains used in their respective offline evaluations.
In particular, the hardware regressor reported in Table~\ref{tab:results_prediction} is not the Lab-only checkpoint that initializes CL-P and Replay in the sequential experiment. Both offline regressors are tested on held-out causally associated visual-contact pairs. 

Table~\ref{tab:results_prediction} reports the mean absolute error (MAE), $R^2$, and the empirical coverage $C_{80}$ of the central 80\% Student's t predictive interval. 
The coverage leverages the entire predictive distribution, including both aleatoric and epistemic components; the closeness to the nominal coverage is interpreted alongside the prediction error and likelihood. 
These measurements evaluate the agreement with proprioceptive interaction indicators, and not the accuracy with respect to intrinsic material constants.

\begin{table}[t]
    \centering
    \caption{Offline prediction on held-out contact-level tests. Within each domain, a separate multi-terrain acquisition dataset was divided into 70\% training, 15\% validation, and 15\% testing. Contacts sharing a timestamp or identical descriptor were grouped. MAE is in the shown units and $C_{80}$ is a percentage (nominal: 80).}
    \label{tab:results_prediction}
    \small
    \setlength{\tabcolsep}{4pt}
    \begin{tabular}{llccc}
        \hline
        Domain & Indicator (units) & MAE & $R^2$ & $C_{80}$ \\
        \hline
        Sim. & Planar foot Slip (m) & 0.0050 & 0.3958 & 82.11\% \\
             & Traction index (--) & 0.0711 & 0.6037 & 74.80\% \\
        \hline
        Real & Planar foot Slip (m) & 0.0047 & 0.4110 & 80.21\%\\
             & Normal GRF (N) & 3.6191 & 0.3470 & 80.21\% \\
             & Traction index (--) & 0.0296 & 0.5269 & 82.81\% \\
             & CoT (--) & 0.235608 & 0.3128 & 85.42\% \\
             & Loading rate (N/s) & 321.605 & 0.5908 & 71.36\% \\
        \hline
    \end{tabular}
\end{table}

To evaluate spatial separation, we associate valid mapped cells with independently annotated surface regions, excluding boundaries with mixed support. 
We measure the score gap between favorable and unfavorable $\Delta\tau=\mathbb{E}[\tau\mid\mathrm{favorable}]-\mathbb{E}[\tau\mid\mathrm{unfavorable}]$, averaging first within each surface and run. The measured gaps are $\Delta\tau_{sim}=0.2991$ and $\Delta\tau_{real}=0.4382$; the smaller simulated gap may reflect the simulated camera's distant terrain view. 
Fig.~\ref{fig:dataflow} illustrates the hardware prediction and its projection into the traversability map.

\subsection{Continual Acquisition and Retention}

We conduct a hardware ablation study to evaluate whether the historical promotion gate improves retention beyond replay-based candidate training.
A regressor trained only on a laboratory-floor acquisition initializes both CL-P and Replay.
Its offline anchor acquisition was excluded from checkpoint training and model selection.
For this study the anchor is held out from gradient updates: both methods replay previously accepted \emph{online} contacts only.
Separate laboratory-floor and new-terrain test acquisitions are used only to report standardized MAE. Contacts from these test acquisitions are excluded from candidate optimization, replay memory, both recent and historical validation sets, and candidate-promotion decisions.

The hardware comparison uses one recorded contact stream with a fixed sequence of artificial grass, textured foam, and foam phases.
It contains $1072$ received observations, of which $951$ contacts are accepted after filtering; the phase ends correspond to $297$, $583$, and $951$ accepted contacts.
The same stream is replayed for both methods with five optimization seeds ($42$--$46$), identical scalers and checkpoint, and the same maximum of $50$ optimizer steps per candidate.
Replay keeps the recent-validation promotion test but disables the anchor constraint; CL-P adds the constraint in Eq.~(\ref{eq:paper_promotion_gate}).

We define new-terrain adaptation as the percentage reduction in standardized MAE on its independent test set from immediately before to immediately after its stream phase, averaging the three terrain gains.
Laboratory-floor degradation is the percentage increase on the independent Lab test from the initial to final deployed model.
The final macro MAE averages the four test-domain MAEs equally.
Anchor NLL degradation uses the same initial-to-final relative change, but on the held-out gate set; it describes the gate mechanism rather than independent generalization.

\begin{table}[t]
    \centering
    \caption{Sequential hardware results, averaged over five optimization seeds on one fixed stream. Lab uses an independent test; anchor NLL uses the gate-validation set, not an independent test. Final MAE is the four-domain macro standardized MAE.}
    \label{tab:hardware_continual}
    \footnotesize
    \setlength{\tabcolsep}{3pt}
    \begin{tabular}{lcccc}
        \hline
        Method & Adapt. $\uparrow$ & Lab $\downarrow$ & Anchor $\downarrow$ & Final MAE $\downarrow$ \\
               & (\%) & (\%) & NLL (\%) & \\
        \hline
        CL-P   & 1.658 & 0.801 & 0.608 & 1.03711 \\
        Replay & 1.664 & 0.924 & 0.790 & 1.03633 \\
        \hline
    \end{tabular}
\end{table}

 \begin{figure}[t]
     \centering
     \includegraphics[width=1.0\linewidth]{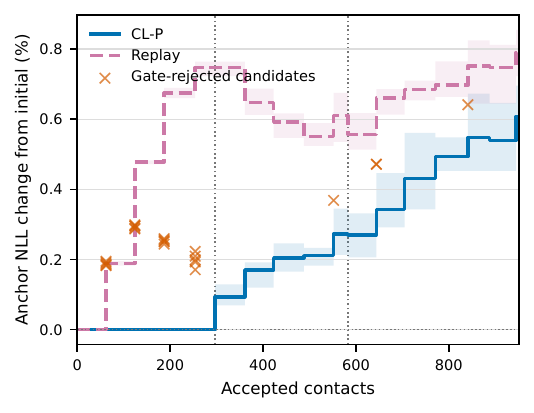}
     \caption{Relative anchor NLL of deployed models during artificial grass $\rightarrow$ textured foam $\rightarrow$ foam. Curves show five-seed means; bands show seed ranges, crosses gate-rejected CL-P candidates, and vertical lines phase transitions. The anchor is used for candidate selection.}
     \label{fig:hardware_anchor_gate}
 \end{figure}

Table~\ref{tab:hardware_continual} shows nearly equal mean new-terrain adaptation for CL-P and Replay ($1.658\%$ versus $1.664\%$).
CL-P initially adapts more slowly to artificial grass ($0.756\%$ versus $1.890\%$), as the gate rejects early candidates, but its gains on textured foam ($3.239\%$ versus $2.329\%$) and foam ($0.979\%$ versus $0.773\%$) compensate.
Both methods continue to improve on previously encountered online terrains in later phases; this short stream does not exhibit a positive error increase on those terrains after their respective phases.

Across $80$ candidate opportunities per method, CL-P promotes $53$ and rejects $24$ specifically by the anchor gate, with three additional recent-validation rejections. Replay promotes $70$, with $10$ recent-validation rejections. 
Figure~\ref{fig:hardware_anchor_gate} shows that rejecting early anchor-degrading candidates keeps the CL-P deployed-model trajectory below Replay.
Mean final anchor NLL degradation is $0.608\%$ for CL-P and $0.790\%$ for Replay, a $23.1\%$ relative reduction.
On the independent Lab test, CL-P also has a slightly smaller mean MAE degradation ($0.801\%$ versus $0.924\%$), although this difference is small across the five seeds.
Replay retains a marginally lower final macro standardized MAE ($1.03633$ versus $1.03711$).
Thus, the gate provides a measurable retention benefit relative to replay alone while largely preserving new-terrain adaptation; it does not establish an overall accuracy advantage on every terrain or across independently acquired hardware streams.

\subsection{Traversability-Aware Navigation}

\begin{figure}[t]
    \centering
    \includegraphics[width=1.0\linewidth]{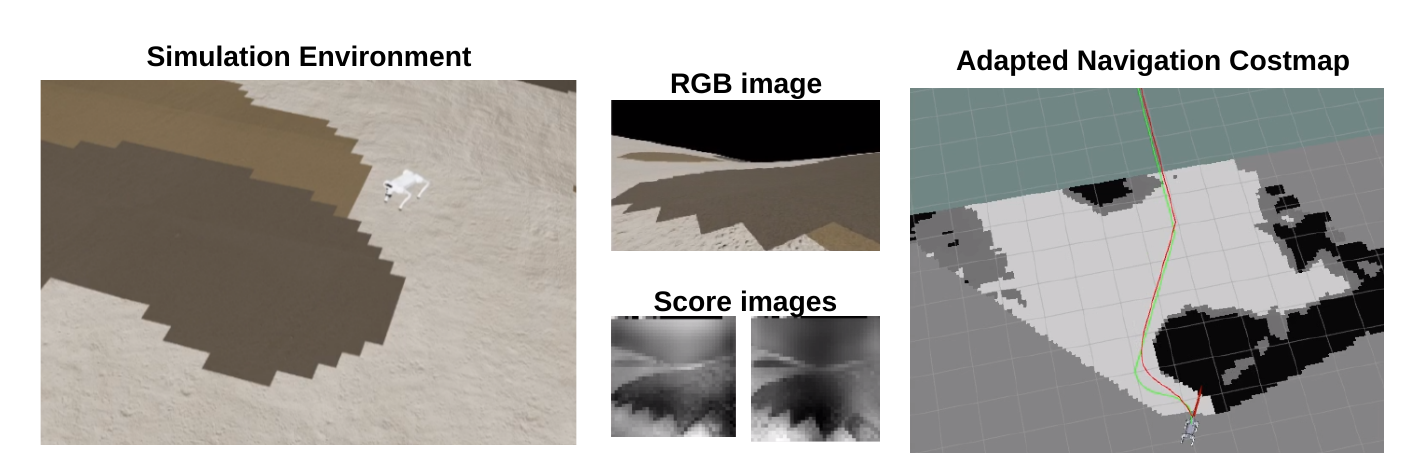}
    \caption{Traversability-aware navigation in simulation: environment and camera view (left), predicted interaction-property scores (center), and resulting navigation costmap and planned path (right).}
    \label{fig:sim_nav}
\end{figure}

We evaluate whether the predicted traversability map can support closed-loop navigation in simulation. Nav2 is used with the SMAC Lattice global planner in the environment described in Sec.~\ref{sec:experimental_setup}, which contains elevation changes and three simulated surfaces with different friction coefficients. These terrains are distinct from those used in the sequential hardware experiment. Fig.~\ref{fig:sim_nav} illustrates the resulting navigation costmap.

We compare Offline and CL-P over five trials with identical initial conditions and a goal $15\,\mathrm{m}$ from the robot. A trial succeeds if the robot reaches the goal within $180\,\mathrm{s}$, with position and yaw tolerances of $0.2\,\mathrm{m}$ and $0.25\,\mathrm{rad}$, without falling or operator intervention. We report success rate, path length $L$, and unfavorable-terrain exposure $E_{\mathrm{bad}}$, defined as the percentage of steps taken in unfavorable regions.

Both methods completed all five trials. Offline and CL-P achieved mean unfavorable-terrain exposures of 3.13\% and 2.76\%, respectively, and mean path lengths of 23.75~m and 24.53~m. The comparable results indicate that CL-P updates preserve downstream navigation performance under the tested conditions. These trials complement the sequential hardware analysis but do not isolate the gate contribution or evaluate navigation after the hardware adaptation stream.

\section{CONCLUSIONS}

This paper presents a quadruped traversability pipeline that predicts five interaction indicators from pre-contact images, adapts through validation-gated replay, and maps property predictions and epistemic uncertainty for navigation.

In a sequential hardware stream, CL-P reduced final anchor NLL degradation by $23.1\%$ relative to Replay with similar new-terrain adaptation. In simulation, Offline and CL-P each completed all five navigation trials, with $3.13\%$ and $2.76\%$ exposure to unfavorable terrain, respectively. These results support guarded adaptation under the tested conditions, but do not establish an overall accuracy advantage or isolate the gate's effect on navigation.

The hardware comparison uses one recorded stream with five optimization seeds, while simulation and hardware use separately trained models. Future work will test longer terrain sequences, deformable terrain, and varying locomotion conditions, as well as uncertainty estimation and active sampling on unfamiliar terrain.

%%%%%%%%%%%%%%%%%%%%%%%%%%%%%%%%%%%%%%%%%%%%%%%%%%%%%%%%%%%%%%%%%%%%%%%%%%%%%%%%%
\bibliographystyle{IEEEtran}
\bibliography{main}

\end{document}